\documentclass[10pt, conference, compsocconf]{IEEEtran}

\ifCLASSINFOpdf
\else
\fi
\usepackage{amsmath}
\usepackage{amssymb}
\usepackage{times}
\usepackage{multicol}
\usepackage{multirow}
\usepackage{bbding}
\usepackage{booktabs}
\usepackage{algorithm}
\usepackage{algorithmic}
\usepackage{graphicx}
\usepackage{epstopdf}
\usepackage{subfigure}
\usepackage{color}
\usepackage{xcolor}
\usepackage[bookmarks=false]{hyperref}
\usepackage{url}
\usepackage{array}
\usepackage{bm}
\usepackage{enumitem}
\usepackage[misc]{ifsym}
\begin{document}
	\title{Accelerating Deep Unsupervised Domain	Adaptation with Transfer Channel Pruning}

\author{\IEEEauthorblockN{
		Chaohui Yu\IEEEauthorrefmark{1}\IEEEauthorrefmark{2},
		Jindong Wang\IEEEauthorrefmark{1}\IEEEauthorrefmark{2},
		Yiqiang Chen\IEEEauthorrefmark{1}\IEEEauthorrefmark{2}\Envelope,
		Zijing Wu\IEEEauthorrefmark{3}
	}
	\IEEEauthorblockA{\IEEEauthorrefmark{1}Beijing Key Lab. of Mobile Computing and Pervasive Device, Inst. of Computing Technology, CAS, Beijing, China}
	\IEEEauthorblockA{\IEEEauthorrefmark{2}University of Chinese Academy of Sciences, Beijing, China}
	\IEEEauthorblockA{\IEEEauthorrefmark{3}Columbia University, New York, NY 10024, USA}
	Email:\{yuchaohui17s,wangjindong,yqchen\}@ict.ac.cn, zw2442@columbia.edu}
	
\maketitle

\begin{abstract}
\label{abstract}
Deep unsupervised domain adaptation~(UDA) has recently received increasing attention from researchers. 
However, existing methods are computationally intensive due to the computation cost of Convolutional Neural Networks~(CNN) adopted by most work. To date, there is no effective network compression method for accelerating these models. In this paper, we propose a unified \textit{Transfer Channel Pruning~(TCP)} approach for accelerating UDA models. 
TCP is capable of compressing the deep UDA model by pruning less important channels while simultaneously learning transferable features by reducing the cross-domain distribution divergence. Therefore, it reduces the impact of negative transfer and maintains competitive performance on the target task. 
To the best of our knowledge, TCP is the \textit{first} approach that aims at accelerating deep UDA models. TCP is validated on two benchmark datasets – Office-31 and ImageCLEF-DA with two common backbone networks – VGG16 and ResNet50. 
Experimental results demonstrate that TCP achieves comparable or better classification accuracy than other comparison methods while significantly reducing the computational cost. 
To be more specific, in VGG16, we get even higher accuracy after pruning 26\% floating point operations (FLOPs); in ResNet50, we also get higher accuracy on half of the tasks after pruning 12\% FLOPs. We hope that TCP will open a new door for future research on accelerating transfer learning models.
\end{abstract}

\IEEEpeerreviewmaketitle

\section{Introduction}
Deep neural networks have significantly improved the performance of diverse machine learning applications. However, in order to avoid 
overfitting and achieve better performance, a large amount of labeled data is needed to train a deep network. Since the manual labeling of massive training data 
is usually expensive in terms of money and time, it is urgent to develop effective algorithms to reduce the labeling workload on the domain to be learned (i.e. \textit{target} domain).

A popular solution to solve the above problem is called \textit{transfer learning}, or \textit{domain adaptation}~\cite{pan2010survey}, which tries to transfer knowledge from well-labeled domains (i.e. \textit{source} domains) to the target domain. 
Specifically, Unsupervised Domain Adaptation (UDA) is considered more challenging since the target domain has no labels. The key is to learn a discriminative model to reduce the distribution divergence between domains. In recent years, deep domain adaptation methods have produced competitive performance in various tasks~\cite{zhao2018finding,long2015learning,venkateswara2017deep}. 
This is because that they take advantages of CNN to learn more transferable representations~\cite{zhao2018finding,long2015learning} compared to traditional methods. 
Popular CNN architectures such as AlexNet~\cite{krizhevsky2012imagenet}, VGGNet~\cite{simonyan2014very}, and ResNet~\cite{he2016deep} are widely adopted as the backbone 
networks for deep unsupervised domain adaptation methods. Then, knowledge can be transferred to the target domain by reducing the cross-domain distance such as Maximum Mean Discrepancy (MMD)~\cite{wang2018stratified} or KL divergence~\cite{pan2011domain}.

Unfortunately, it is still challenging to deploy these deep UDA models on resource constrained devices such as mobile phones since there is a huge computational cost required by these methods. In order to reduce resource requirement and accelerate the inference process, a common solution is \textit{network compression}. 
Network compression methods mainly include network quantization~\cite{rastegari2016xnor,zhou2017incremental}, weight pruning \cite{han2015learning,he2017channel,molchanov2016pruning}, and low-rank approximation \cite{denton2014exploiting,han2015deep}. 
Especially channel pruning~\cite{he2017channel,molchanov2016pruning,yu2017accelerating}, which is a type of weight pruning and compared to other methods, it does not need special hardware or software implementations. 
And it can reduce \textit{negative transfer}~\cite{pan2010survey} by pruning some redundant channels, in which, negative transfer refers to the knowledge learned on the source domain that has a negative effect on the learning on the target domain. So it is a good choice for compressing deep UDA models.

However, it is not feasible to apply the above network compression methods directly to the UDA problems. The reasons are two folds. Firstly, these compression methods are proposed to solve \textit{supervised} learning problems, which is not suitable for the UDA settings since there are no labels in the target domain. Secondly, even if we can acquire some labels manually, applying these compression methods directly to UDA will result in negative transfer, since they fail to consider the distribution discrepancy between the source and target domains. Currently, there is no effective network compression method for UDA.

\begin{figure*}[t!]
	\centering\includegraphics[scale=0.5]{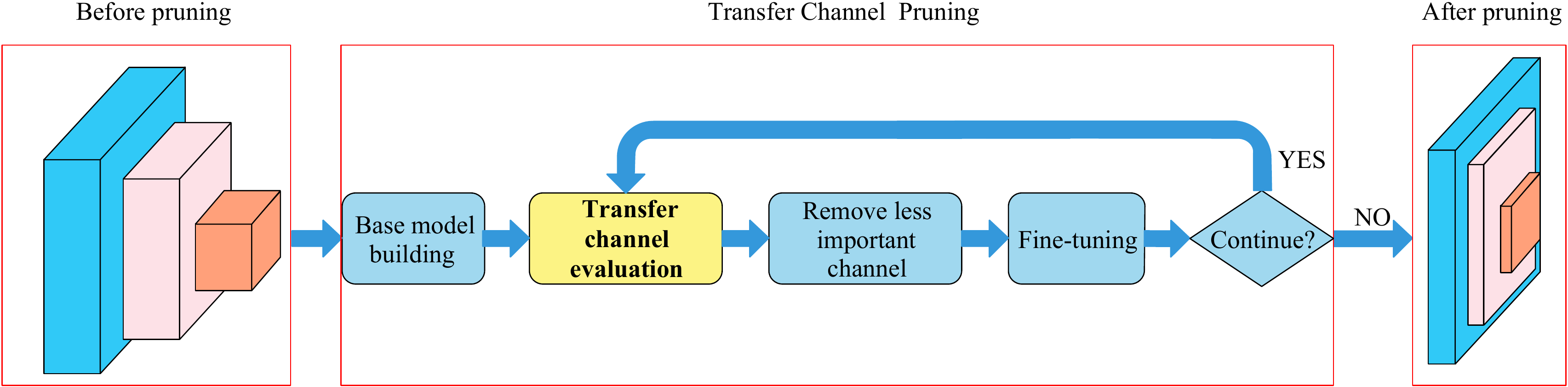}
	\vspace{-.1in}
	\caption{The framework of the proposed Transfer Channel Pruning (TCP) approach.}
	\label{fig-framework}
	\vspace{-.1in}
\end{figure*}

In this paper, we propose a unified network compression method called \textbf{Transfer Channel Pruning~(TCP)} for accelerating deep unsupervised domain adaptation models. The general framework of our method TCP is shown in Fig.~\ref{fig-framework}. 
Starting from a deep unsupervised domain adaptation base model, TCP iteratively evaluates the importance of channels with the transfer channel evaluation module and remove less important channels for both source and target domains. 
TCP is capable of compressing the deep UDA model by pruning less important channels while simultaneously learning transferable features by reducing the cross-domain distribution divergence. 
Experimental results demonstrate that TCP achieves better classification accuracy than other comparison pruning methods while significantly reducing the computational cost. To the best of our knowledge, TCP is the \textit{first} approach to accelerate the deep UDA models.

To summarize, the contributions of this paper are as follows: 

1)~We present TCP as a unified approach for accelerating deep unsupervised domain adaptation models. TCP is a generic, accurate, and efficient compression method that can be easily implemented by most deep learning libraries.

2)~TCP is able to reduce negative transfer by considering the cross-domain distribution discrepancy using the proposed \textit{Transfer Channel Evaluation} module.

3)~Extensive experiments on two public UDA datasets demonstrate the significant superiority of our TCP method.


\section{Related Work}
\label{sec-related}
Our work is mainly related to unsupervised domain adaptation and network compression.

\subsection{Unsupervised Domain Adaptation}
UDA is a specific area of transfer learning~\cite{pan2010survey}, which 
is to learn a discriminative model in the presence of the domain-shifts between domains. 
The main problem of UDA is how to reduce the domain shift between the source and target domains. 
There are many methods to tackle this problem: traditional (shallow) learning and deep learning. 

Traditional (shallow) learning methods have several aspects: 
1) Subspace learning. Subspace Alignment (SA) \cite{fernando2013unsupervised} aligns the base vectors of both domains and Subspace Distribution Alignment (SDA) 
\cite{sun2015subspace} extends SA by adding the subspace variance adaptation. 
CORAL \cite{sun2016return} aligns subspaces in second-order statistics. 
2) Distribution alignment. Joint Distribution Adaptation (JDA) \cite{long2013transfer} is proposed to match both distributions with equal weights. Later works extend JDA by adding structural consistency \cite{hou2016unsupervised} and domain invariant clustering \cite{tahmoresnezhad2017visual}. 
But these works treat the two distributions equally and fail to leverage the different importance of distributions. Recently, Wang \textit{et al.} proposed the Manifold Embedded Distribution Alignment~(MEDA)~\cite{wang2018visual,wang2017balanced} approach to dynamically evaluate the different effect of marginal and conditional distributions and achieved the state-of-the-art results on domain adaptation.

As for deep learning methods, CNN can learn nonlinear deep representations and capture underlying factors of variation between different tasks \cite{bengio2013representation}. 
These deep representations can disentangle the factors of variation, which enables the transfer of knowledge between tasks \cite{oquab2014learning}. 
Recent works on deep UDA embed domain-adaptation modules into deep networks to improve transfer performance \cite{glorot2011domain,kumar2018co,ganin2016domain,tzeng2017adversarial}, where significant performance gains have been obtained. UDA has wide applications in computer vision \cite{long2015learning,hoffman2017cycada} and natural language processing \cite{zhao2018finding} and is receiving increasing attention from researchers.

As far as we know, no previous UDA approach has focused on the acceleration of the network.

\subsection{Network Compression}
These years, for better accuracy, designing deeper and wider CNN models has become a general trend, such 
as VGGNet \cite{simonyan2014very} and ResNet \cite{he2016deep}. However, as the CNN grow bigger, it is harder to deploy these deep models on resource constrained devices. Network compression becomes an efficient way to solve this problem. Network compression methods mainly include network quantization, low-rank approximation and weight pruning. Network quantization is good at decreasing the presentation precision of parameters so as to reduce the storage space. Low-rank approximation reduces the storage space by low-rank matrix techniques, which is not efficient for point-wise convolution~\cite{chollet2017xception}. Weight pruning mainly includes two methods, neural pruning~\cite{han2015learning,lin2017runtime} and channel pruning~\cite{he2017channel,molchanov2016pruning,yu2017accelerating,luo2017thinet}. 

Channel pruning methods prune the whole channel each time, so it is fast and efficient than neural pruning which removes a single neuron connection each time. It is a structured pruning method, compared to network quantization and low-rank approximation, it does not introduce sparsity to the original network structure and also does not require special software or hardware implementations. It has demonstrated superior performance compared to other methods and many works~\cite{he2017channel,molchanov2016pruning,luo2017thinet} have been proposed to perform channel pruning on pre-trained models with different kinds of criteria.

These above pruning methods mainly aim at supervised learning problems, by contrast, there have been few studies for compressing unsupervised domain adaptation models. 
As far as we know, we are the first to study how to do channel pruning for deep unsupervised domain adaptation.

TCP is primarily motivated by \cite{molchanov2016pruning}, while our work is different from it. TCP is presented for pruning unsupervised domain adaptation models. 
To be more specific, we take the discrepancy between the source and target domains into consideration so we can prune the less important channels not just for the source domain but also for the unlabeled target domain. We call this Transfer Channel Evaluation, which is highlighted in yellow in Fig.~\ref{fig-framework}.

\section{Transfer Channel Pruning}
\label{sec-method}
In this section, we introduce the proposed Transfer Channel Pruning (TCP) approach. 

\subsection{Problem Definition}
\label{sec-problem}

In unsupervised domain adaptation, we are given a source domain $\mathcal{D}_{s}=\{(\mathbf{x}^{s}_{i},y^{s}_{i})\}^{n_s}_{i=1}$ of $n_{s}$ labeled examples and 
a target domain $\mathcal{D}_{t}=\{\mathbf{x}^{t}_{j}\}^{n_t}_{j=1}$ of $n_{t}$ unlabeled examples. $\mathcal{D}_{s}$ and $\mathcal{D}_{t}$ have the same label space, i.e. $\mathbf{x}_{i}, \mathbf{x}_{j} \in \mathbb{R}^d$ where $d$ is the dimensionality. The marginal distributions between two domains are different, i.e. $P_s(\mathbf{x}_s) \ne P_t(\mathbf{x}_t)$. The goal of deep UDA is to design a deep neural network that enables learning of transfer classifiers $y~=~f_{s}(\mathbf{x})$ and $y~=~f_{t}(\mathbf{x})$ to close the source-target discrepancy and can achieve the best performance on the target dataset. 

For a pre-trained deep UDA model, its parameters can be denoted as $\mathbf{W}$. Here we assume the $l_{th}$ convolutional layer has an output activation tensor $\mathbf{a}_{l}$ of size of $h_{l} \times w_{l} \times k_{l}$, where $k_{l}$ represents the number of output channels of the $l_{th}$ layer, and $h_{l}$ and $w_{l}$ stand for the height and width of feature maps of the $l_{th}$ layer, respectively. Therefore, the goal of TCP is to prune a UDA model in order to accelerate it with comparable or even better performance on the target domain. In this way, we can obtain smaller models that require less computation complexity and memory consumption, which can be deployed on resource constrained devices.

\subsection{Motivation}
\label{sec-mot}

We compress the deep UDA model using model pruning methods for their efficiency. A straightforward model pruning technique is a \textit{two-stage} method, which first prunes the model on the source domain with supervised learning and then fine-tunes the model on the target domain. However, negative transfer~\cite{pan2010survey} is likely to happen during this pruning process since the discrepancy between the source and target domains is ignored.

In this work, we propose a unified \textbf{Transfer Channel Pruning~(TCP)} approach to tackle such challenge. TCP is capable of compressing the deep UDA model by pruning less important channels while simultaneously learning transferable features by reducing the cross-domain distribution divergence. Therefore, TCP reduces the impact of negative transfer and maintains competitive performance on the target task. In short, TCP is a generic, accurate, and efficient compression method that can be easily implemented by most deep learning libraries.

To be more specific, Fig.~\ref{fig-framework} illustrates the main idea of TCP. There are mainly three steps. Firstly, TCP learns the base deep UDA model through \textit{Base Model Building}. The base model is fine-tuned with the standard UDA criteria. Secondly, TCP evaluates the importance of channels of all layers with the \textit{Transfer Channel Evaluation} and performs further fine-tuning. Specifically, the convolutional layers, which usually dominate the
computation complexity, are pruned in this step. Thirdly, TCP \textit{iteratively refines} the pruning results and stops after reaching the trade-off between accuracy and FLOPs (i.e. computational cost) or parameter size.

\begin{figure}[t!] 
	\centering 
	\includegraphics[scale=0.4]{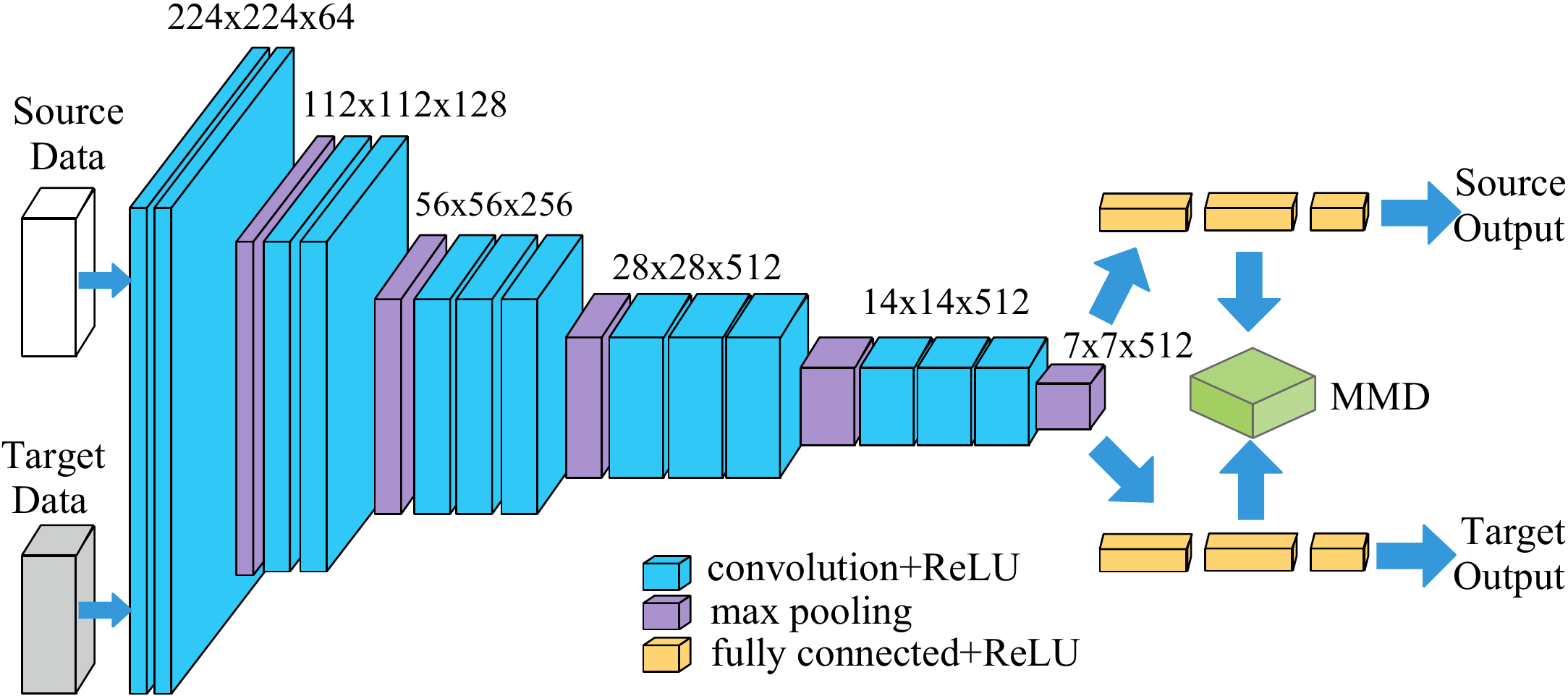}
	\vspace{-.1in}
	\caption{The basic deep UDA architecture.}
	\label{fig-dan}
	\vspace{-.1in}
\end{figure}

\subsection{Base Model Building}
\label{sec-danbase}
In this step, we build the base UDA model with deep neural networks. Deep neural networks have been successfully used in UDA with state-of-the-art algorithms \cite{ganin2016domain,long2015learning,tzeng2017adversarial} in recent years. Previous studies~\cite{yosinski2014transferable,tzeng2017adversarial} have shown that the features extracted by deep networks are general at lower layers, while specific at the higher layers since they are more task-specific. Therefore, more transferable representations can be learned by transferring the features at lower layers and then fine-tune the task-specific layers. During fine-tuning, the cross-domain discrepancy can be reduced by certain adaptation distance. Since our main contribution is not designing new deep UDA networks, we adopt DAN~\cite{long2015learning} as the base architecture. DAN is a popular deep UDA approach and its variants have been widely adopted for UDA tasks. In the following, we will briefly introduce the main idea of DAN, and more details can be found in its original paper.

As shown in Fig.~\ref{fig-dan}, we learn transferable features via several convolutional and pooling layers~(the blue and purple blocks). Then, the classification task can be accomplished with the fully-connected layers~(the yellow blocks). Maximum Mean Discrepancy (MMD)~\cite{pan2011domain} is adopted as the adaptation loss in order to reduce domain shift. MMD has been proposed to provide the distribution difference between the source and target datasets. And it has been widely utilized in many UDA methods~\cite{long2015learning,pan2008transfer,tzeng2017adversarial}. 
The MMD loss between two domains can be computed as 
\begin{equation}
	\label{equ-mmd}
	L_{mmd}=
	\begin{Vmatrix}
	\frac{1}{n_s} \sum \limits_{\mathbf{x}_i \in \mathcal{D}_s}^{} \phi(\mathbf{x}_{i}) - \frac{1}{n_t}\sum \limits _{\mathbf{x}_j \in \mathcal{D}_t}^{} \phi(\mathbf{x}_{j})
	\end{Vmatrix}^2_{\mathcal{H}},
\end{equation}
where $\mathcal{H}$ denotes Reproducing Kernel Hilbert Space~(RKHS) with gaussian kernel and $\phi(\cdot)$ denotes some feature map to map the original samples to RKHS. 

Several popular architectures can serve as the backbone network of DAN, such as AlexNet~\cite{krizhevsky2012imagenet}, VGGNet~\cite{simonyan2014very}, and ResNet~\cite{he2016deep}. After obtaining the base model, we can perform channel pruning to accelerate the model.

\subsection{Transfer Channel Evaluation}
\label{sec-tcp}
The goal of transfer channel evaluation is to iteratively evaluate the importance of output channels of layers in order to prune the $\mathcal{K}$ least important channels. Here $\mathcal{K}$ is controlled by users. In the pruning process, we want to preserve and refine a set of parameters $\mathbf{W}^{\prime}$, which represents those
important parameters for both source and target domains. 
Let $L(\mathcal{D}_{s},\mathcal{D}_{t},\mathbf{W})$ be the cost function for UDA and $\mathbf{W^{\prime}}=\mathbf{W}$ at the starting time. 
For a better set of parameters $\mathbf{W}^{\prime}$, we want to minimize the loss change after pruning a channel $\mathbf{a}_{l,i}$. This can be considered as an optimization problem. Here we introduce the absolute difference of loss:
\begin{equation}
	\label{equ-Ldis}
	|\Delta L(\mathbf{a}_{l,i})| = |L(\mathcal{D}_{s},\mathcal{D}_{t},\mathbf{a}_{l,i}) - L(\mathcal{D}_{s},\mathcal{D}_{t},\mathbf{a}_{l,i} = 0)|,
\end{equation}
which means the loss change after pruning the $i_{th}$ channel of the $l_{th}$ convolutional layer. And we want to minimize $|\Delta L(\mathbf{a}_{l,i})|$ by selecting the 
appropriate channel $\mathbf{a}_{l,i}$. Pruning will stop until a trade-off between accuracy and pruning 
object (FLOPs or parameter size) has been achieved.

However, it is hard to find a set of optimal parameters in one go, because the search space is $2^{|\mathbf{W}|}$
which is too huge to compute and try every combination. Inspired by \cite{molchanov2016pruning}, our TCP solves this problem with a greedy algorithm 
by iteratively removing the $\mathcal{K}$ least important channels at each time.

\subsubsection{\textbf{Criteria}}
Criteria is the criterion for judging the importance of channels. 
Since the key to channel pruning is to select the least important channel, especially for UDA, we design the criteria of TCP carefully. There are many heuristic criteria, including the $L_{2}$-norm of filter weights, the activation statistics of feature maps, mutual information between activations and predictions and Taylor expansion, etc. Here we choose the \textit{first-order Taylor expansion} as the base criteria since its efficiency and performance has been verified in~\cite{molchanov2016pruning} for pruning supervised learning models. Compared with our TCP, we also take pruning as an optimization problem, however, the objective we want to optimize is the final performance on the unlabeled target dataset. So we design our criteria in a different way which is better for pruning deep UDA models.

According to Taylor's theorem, the Taylor expansion at point $x = a$ can be computed as:
\begin{equation}
	\label{equ-taylor}
	f(x) = \sum_{p=0}^{P} \frac{f^{(p)}(a)}{p!}(x-a)^{p} + R_{p}(x),
\end{equation}
where $p$ denotes the $p_{th}$ derivative of $f(x)$ at point $x = a$ and the last item $R_{p}(x)$ represents the $p_{th}$ remainder. To approximate 
$|\Delta L(\mathbf{a}_{l,i})|$, we can use the first-order Taylor expansion near $\mathbf{a}_{l,i} = 0$ which means the loss change after removing $\mathbf{a}_{l,i}$, then we can get:
\begin{equation}
	\label{equ-firsttaylor}
	f(\mathbf{a}_{l,i}=0) = f(\mathbf{a}_{l,i}) - f^{\prime}(\mathbf{a}_{l,i})\cdot\mathbf{a}_{l,i} + \frac{|\mathbf{a}_{l,i}|^{2}}{2} \cdot f^{\prime\prime}(\xi),
\end{equation}
where $\xi$ is a value between 0 and $\mathbf{a}_{l,i}$, and $\frac{|\mathbf{a}_{l,i}|^{2}}{2} \cdot f^{\prime\prime}(\xi)$ is a Lagrange form remainder which 
requires too much computation, so we abandon this item for accelerating the pruning process. Then back to Eq.~(\ref{equ-Ldis}), we can get:
\begin{equation}
	\label{equ-ali}
	L(\mathcal{D}_{s},\mathcal{D}_{t},\mathbf{a}_{l,i} = 0)	= L(\mathcal{D}_{s},\mathcal{D}_{t},\mathbf{a}_{l,i}) - \frac{\partial{L}}{\partial{\mathbf{a}_{l,i}}}\cdot\mathbf{a}_{l,i}.
\end{equation}

Then, we combine Eq.~(\ref{equ-Ldis}) and Eq.~(\ref{equ-ali}) and get the criteria $G$ of TCP:
\begin{equation}
	\label{equ-deltaL}
	G(\mathbf{a}_{l,i}) = |\Delta L(\mathbf{a}_{l,i})| = |\frac{\partial{L}}{\partial{\mathbf{a}_{l,i}}}\cdot\mathbf{a}_{l,i}|,
\end{equation}
which means the absolute value of product of the activation and the gradient of the cost function, and $\mathbf{a}_{l,i}$ can be calculated as:
\begin{equation}
	\label{equ-activation}
	\mathbf{a}_{l,i} = \frac{1}{N} \sum_{n=1}^{N} \frac{1}{h_{l} \times w_{l}}\sum_{p=1}^{h_{l}}\sum_{q=1}^{w_{l}}\mathbf{a}^{p,q}_{l,i},
\end{equation}
where $N$ is the number of batch size and $\mathbf{a}^{p,q}_{l,i}$ is the value of the $p_{th}$ row and the $q_{th}$ column of the activated feature map $\mathbf{a}_{l,i}$. 

\subsubsection{\textbf{Loss Function of TCP}}
To make TCP focus on pruning UDA models, we simultaneously take the source domain and the unlabeled target domain into consideration. 
The loss function of TCP consists of two parts, $L_{cls}(\mathcal{D}_{s},\mathbf{W})$ and $L_{mmd}(\mathcal{D}_{s},\mathcal{D}_{t},\mathbf{W})$. Here, 
$L_{cls}(\mathcal{D}_{s},\mathbf{W})$ is a cross-entropy loss which denotes the classification loss on source domain and can be computed as:
\begin{equation}
	\label{equ-CEloss}
	L_{cls}(\mathcal{D}_{s},\mathbf{W})~=~-\frac{1}{N} \sum_{i=1}^{N} \sum_{c=1}^{C} P_{i,c} log(\textit{h}_{c}(x^{s}_{i}))
\end{equation}
where $C$ is the number of classes of source dataset, $P_{i,c}$ is the probability of $x^{s}_{i}$ belonging to class $c$, and $\textit{h}_{c}(x^{s}_{i})$ denotes the probability that the model predicts $x^{s}_{i}$ as class $c$. And $L_{mmd}(\mathcal{D}_{s},\mathcal{D}_{t},\mathbf{W})$ denotes the MMD loss between the source and target domains that presented in Eq.~(\ref{equ-mmd}). The total loss function can be computed as:
\begin{equation}
	\label{equ-loss}
	\begin{split}
	L(\mathcal{D}_{s},\mathcal{D}_{t},\mathbf{W})~=~&L_{cls}(\mathcal{D}_{s},\mathbf{W})~+ \\
	~& \beta L_{mmd}(\mathcal{D}_{s},\mathcal{D}_{t},\mathbf{W}),
	\end{split}
\end{equation}
where
\begin{equation}
	\label{equ-beta}
	\beta~=~\frac{4}{1+e^{-1\cdot \frac{i}{ITER}}} - 2.
\end{equation}

Here, $\beta$ is a dynamic value which takes values in $(0,1)$. $i \in (0,{ITER})$ where $ITER$ is the 
number of pruning iterations. We design $\beta$ in this way for two main reasons, on the one hand, during the early stage of pruning, the weights have not converged 
and keep unstable so the $L_{mmd}$ is too large and makes the pruned model hard to converge. On the other hand, in the rest of the pruning process, the $L_{mmd}$ becomes more important that can guide the pruned model to focus more on the target domain. So the criteria of TCP can be computed as: 
\begin{equation}
	\label{equ-criteria}
	\begin{split}
	G(\mathbf{a}_{l,i})~=~&|\frac{\partial{L_{cls}(\mathcal{D}_{s},\mathbf{W})}}{\partial\mathbf{a}^{s}_{l,i}}\cdot\mathbf{a}^{s}_{l,i}~+ \\
	~& \beta \frac{\partial{L_{mmd}(\mathcal{D}_{s},\mathcal{D}_{t},\mathbf{W})}}{\partial\mathbf{a}^{t}_{l,i}}\cdot\mathbf{a}^{t}_{l,i}|,
	\end{split}
\end{equation} 
where $\mathbf{a}^{s}_{l,i}$ and $\mathbf{a}^{t}_{l,i}$ denote the activation with source data and target data respectively.

\subsection{Iterative Refinement}
\label{sec-postp}
After the transfer channel evaluation, each channel is sorted according to Eq.~(\ref{equ-criteria}) and the $\mathcal{K}$ least important channels are removed after each pruning iteration. Then, a short-term fine-tuning is adopted to the pruned model to help the model to converge and the pruning is done after a trade-off between accuracy and FLOPs or parameter size has been achieved. In which, the trade-off means both the computational complexity of the model and accuracy on the target domain are all acceptable. And since the target domain has no labels, so in practice, a small target domain test dataset is built by acquiring some labels manually. 
The learning procedure of TCP is described in Algorithm~\ref{algo-tcp}.

\begin{algorithm}[t!]
	\caption{TCP:~\underline{T}ransfer \underline{C}hannel \underline{P}runing}
	\label{algo-tcp}
	\renewcommand{\algorithmicrequire}{\textbf{Input:}} 
	\renewcommand{\algorithmicensure}{\textbf{Output:}} 
	\begin{algorithmic}[1]
		\REQUIRE
		Source domain $\mathcal{D}_{s} = \{(\mathbf{x}^{s}_{i},y^{s}_{i})\}^{n_s}_{i=1}$, 
		target domain $\mathcal{D}_{t} = \{\mathbf{x}^{t}_{j}\}^{n_t}_{j=1}$, the baseline $\mathbf{W}$.\\
		\ENSURE
		A pruned model $\mathbf{W^{\prime}}$ for deep unsupervised domain adaptation.\\
		\STATE Fine-tune the unsupervised domain adaptation baseline until the best performance achieved on the unlabeled target dataset;
		\FOR{$iteration ~ i$}
		\STATE Sort the importance of channels by criteria Eq.~(\ref{equ-criteria}) and identify less significant channels;\
		\STATE Remove the $\mathcal{K}$ least important channels of the layers;\
		\STATE Short-term fine-tune;
		\IF{the trade-off between accuracy on the target domain and FLOPs or parameter size has achieved}
		\STATE $break$
		\ENDIF 
		\ENDFOR
		\STATE Long-term fine-tune;
		\RETURN pruned model $\mathbf{W^{\prime}}$.
	\end{algorithmic}
\end{algorithm}

\section{Experimental Analysis}
\label{sec-exp}
In this section, we evaluate the performance of TCP via experiments on pruning deep unsupervised domain adaptation models. 
We evaluate our approaches for VGGNet \cite{simonyan2014very} and ResNet \cite{he2016deep} on two popular 
datasets – Office-31 \cite{saenko2010adapting} and ImageCLEF-DA~\footnote{\url{http://imageclef.org/2014/adaptation}}.
All our methods are implemented based on the PyTorch~\cite{paszke2017automatic} framework and the code will be released at \url{github.com/jindongwang/transferlearning/code/deep/TCP}.

\subsection{Datasets}
\subsubsection{Office-31} This dataset is a standard and maybe the most popular benchmark for unsupervised domain adaptation. It consists of 4,110 images
within 31 categories collected from everyday objects in an office environment. It consists of three domains: $Amazon$ (\textbf{A}), which contains 
images downloaded from \url{amazon.com}, $Webcam$ (\textbf{W}) and $DSLR$ (\textbf{D}), which contain images respectively taken by web camera and 
digital SLR camera under different settings. A sample of the Office-31 dataset is shown in the left part of Fig.~\ref{fig-office_clef}. 
We evaluate all our methods across six transfer tasks on all the three domains \textbf{A}$\to$\textbf{W}, \textbf{W}$\to$\textbf{A}, \textbf{A}$\to$\textbf{D}, 
\textbf{D}$\to$\textbf{A}, \textbf{D}$\to$\textbf{W} and \textbf{W}$\to$\textbf{D}.

\subsubsection{ImageCLEF-DA} 
\begin{figure}[t!] 
	\centering 
	\includegraphics[scale=0.43]{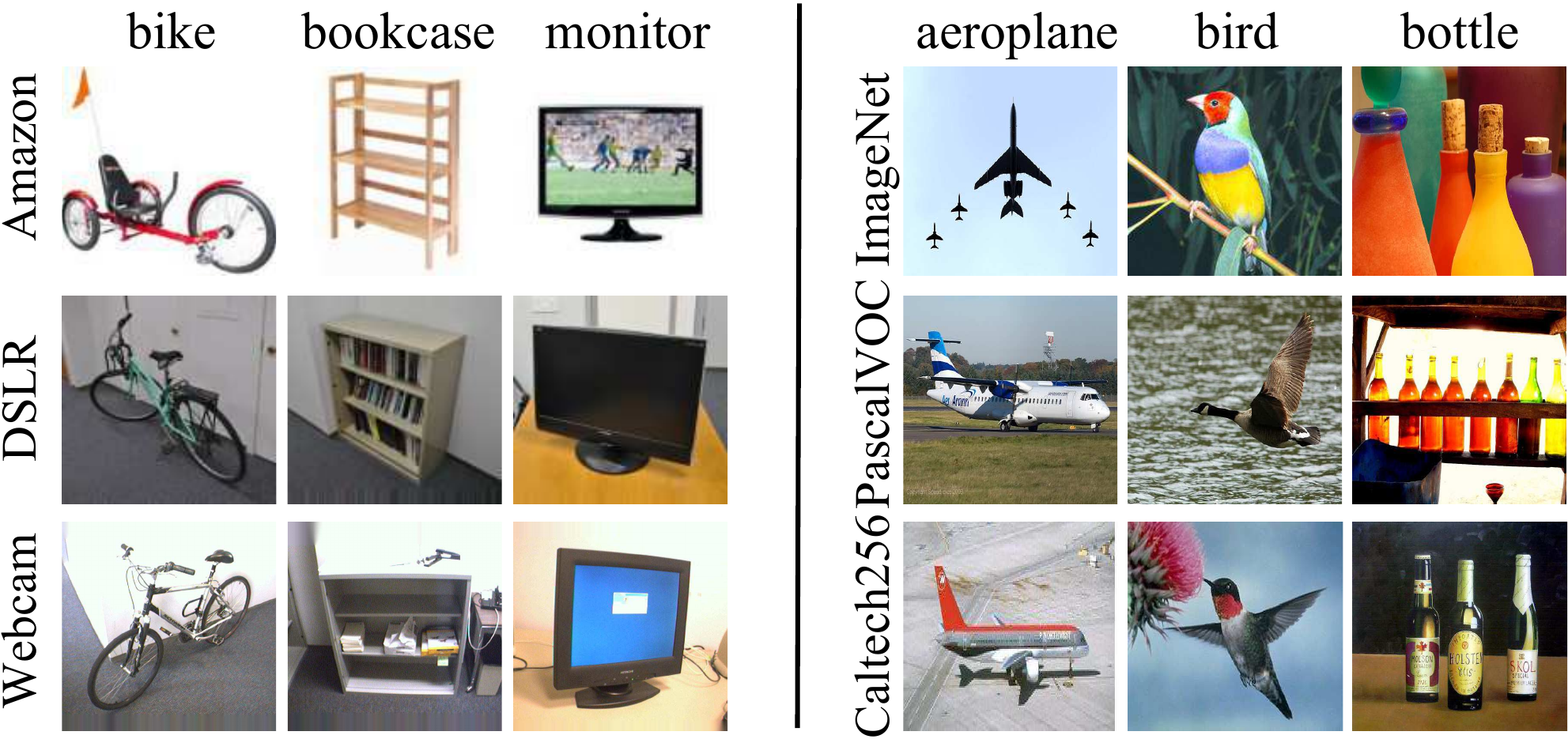}
	\vspace{-.1in}
	\caption{A sample of the two benchmark datasets. Left: the Office-31 dataset; Right: the ImageCLEF-DA dataset.}
	\label{fig-office_clef}
	\vspace{-.2in}
\end{figure}

This dataset is a benchmark dataset for ImageCLEF 2014 domain adaptation challenge, and it is collected by 
selecting the 12 common categories shared by the following public datasets and each of them is considered as a domain: $Caltech-256$ (\textbf{C}), 
$ImageNet~ILSVRC~2012$ (\textbf{I}), $Pascal~VOC~2012$ (\textbf{P}) and $Bing$ (\textbf{B}). There are 50 images in each category and 600 images in each domain. A sample of the ImageCLEF-DA dataset is shown in the right part of Fig.~\ref{fig-office_clef}. We evaluate all methods across six transfer tasks following \cite{long2015learning}: \textbf{I}$\to$\textbf{P}, \textbf{P}$\to$\textbf{I}, \textbf{I}$\to$\textbf{C}, \textbf{C}$\to$\textbf{I}, \textbf{P}$\to$\textbf{C} and \textbf{C}$\to$\textbf{P}. Compared with Office-31, this dataset is more balanced and can be a good comparable dataset to Office-31.

\begin{table*}[t!]
	\centering
	\caption{\upshape The performance on Office-31 dataset (VGG16-based and ResNet50-based). 
	Here, $FLOPs\downarrow$ and $Param\downarrow$ denote the decrement of FLOPs and parameter size compared with the baseline, \textit{Acc} means the accuracy on target domain.}
	\label{tb-office}
	\resizebox{\textwidth}{!}{%
	\begin{tabular}{c|c|cc|cc|cc|cc|cc|cc|cc}
	\hline
	\multirow{2}{*}{Models}  & \multirow{2}{*}{FLOPs$\downarrow$} & \multicolumn{2}{c|}{A$\to$W}          & \multicolumn{2}{c|}{D$\to$W}          & \multicolumn{2}{c|}{W$\to$D}          & \multicolumn{2}{c|}{A$\to$D}          & \multicolumn{2}{c|}{D$\to$A}          & \multicolumn{2}{c|}{W$\to$A}          & \multicolumn{2}{c}{Average}           \\ \cline{3-16} 
							 &                        & Acc             & Param$\downarrow$           & Acc             & Param$\downarrow$           & Acc             & Param$\downarrow$           & Acc             & Param$\downarrow$           & Acc             & Param$\downarrow$           & Acc             & Param$\downarrow$           & Acc             & Param$\downarrow$           \\ \hline
	VGG16-base    & —                      & 74.0\%          & —               & 94.0\%          & —               & 97.5\%          & —               & 72.3\%          & —               & 54.1\%          & —               & 55.2\%          & —               & 74.5\%          & —               \\
	Two\_stage               & \multirow{3}{*}{26\%}  & 69.4\%          & 29.4\%          & 94.5\%          & 31.2\%          & 99.0\%          & 30.6\%          & 69.8\%          & 29.3\%          & 42.7\%          & 32.5\%          & 47.2\%          & 28.3\%          & 70.4\%          & 30.2\%          \\
	TCP\_w/o\_DA             &                        & 73.0\%          & 29.7\%          & 95.5\%          & 32.7\%          & 99.3\%          & 29.2\%          & 75.8\%          & 25.8\%          & 45.9\%          & 29.3\%          & 50.4\%          & 29.6\%          & 73.3\%          & 29.4\%          \\
	TCP                      &                        & \textbf{76.1\%} & \textbf{36.8\%} & \textbf{96.1\%} & \textbf{36.2\%} & \textbf{99.8\%} & \textbf{32.6\%} & \textbf{76.2\%} & \textbf{35.9\%} & \textbf{47.9\%} & \textbf{37.1\%} & \textbf{51.2\%} & \textbf{39.8\%} & \textbf{74.5\%} & \textbf{36.4\%} \\ \hline
	Two\_stage               & \multirow{3}{*}{70\%}  & 57.1\%          & 63.3\%          & 88.1\%          & 68.0\%          & 96.3\%          & 64.5\%          & 55.0\%          & 61.0\%          & 31.8\%          & 66.2\%          & 32.7\%          & 63.1\%          & 60.2\%          & 64.3\%          \\
	TCP\_w/o\_DA             &                        & 53.5\%          & 62.5\%          & 89.2\%          & 61.3\%          & 97.9\%          & 57.5\%          & 61.8\%          & 54.8\%          & 35.3\%          & 62.6\%          & 34.5\%          & 58.8\%          & 62.0\%          & 59.6\%          \\
	TCP                      &                        & \textbf{74.1\%} & \textbf{69.3\%} & \textbf{89.5\%} & \textbf{69.8\%} & \textbf{98.8\%} & \textbf{65.2\%} & \textbf{65.9\%} & \textbf{66.2\%} & \textbf{35.6\%} & \textbf{68.5\%} & \textbf{38.5\%} & \textbf{68.2\%} & \textbf{67.1\%} & \textbf{67.9\%} \\ \hline\hline
	ResNet50-base & —                      & 80.3\%          & —               & 97.1\%          & —               & 99.2\%          & —               & 78.9\%          & —               & 64.3\%          & —               & 62.3\%          & —               & 80.3\%          & —               \\
	Two\_stage               & \multirow{3}{*}{12\%}  & 75.8\%          & 32.4\%          & 96.7\%          & 31.4\%          & 99.5\%          & 35.5\%          & 76.0\%          & 30.4\%          & 48.0\%          & 28.3\%          & 50.1\%          & 29.4\%          & 74.4\%          & 31.2\%          \\
	TCP\_w/o\_DA             &                        & 79.8\%          & 33.5\%          & 97.0\%          & 35.5\%          & \textbf{100\%}  & 34.5\%          & 77.1\%          & 36.2\%          & 47.8\%          & 34.5\%          & 52.6\%          & 33.1\%          & 75.7\%          & 34.5\%          \\
	TCP                      &                        & \textbf{81.8\%} & \textbf{37.7\%} & \textbf{98.2\%} & \textbf{36.2\%} & 99.8\%          & \textbf{37.0\%} & \textbf{77.9\%} & \textbf{36.9\%} & \textbf{50.0\%} & \textbf{35.0\%} & \textbf{55.5\%} & \textbf{36.9\%} & \textbf{77.2\%} & \textbf{36.7\%} \\ \hline
	Two\_stage               & \multirow{3}{*}{46\%}  & 65.5\%          & 56.2\%          & 93.0\%          & 56.3\%          & 98.7\%          & \textbf{57.3\%} & 64.9\%          & 56.0\%          & 34.0\%          & 57.2\%          & 38.9\%          & 57.3\%          & 65.8\%          & 56.7\%          \\
	TCP\_w/o\_DA             &                        & 75.1\%          & 56.4\%          & 95.8\%          & 56.4\%          & 99.2\%          & 56.6\%          & 70.8\%          & 55.8\%          & 34.2\%          & 56.4\%          & 41.5\%          & 56.6\%          & 69.4\%          & 56.4\%          \\
	TCP                      &                        & \textbf{77.4\%} & \textbf{58.4\%} & \textbf{96.3\%} & \textbf{58.0\%} & \textbf{100\%}  & 57.1\%          & \textbf{72.0\%} & \textbf{59.0\%} & \textbf{36.1\%} & \textbf{57.8\%} & \textbf{46.3\%} & \textbf{58.5\%} & \textbf{71.3\%} & \textbf{58.1\%} \\ \hline
	\end{tabular}%
	}
\end{table*}

\begin{table*}[t!]
	\centering 
	\caption{\upshape The performance on ImageCLEF-DA dataset (VGG16-based and ResNet50-based).}
	\label{tb-clef}
	\resizebox{\textwidth}{!}{%
	\begin{tabular}{c|c|cc|cc|cc|cc|cc|cc|cc}
	\hline
	\multirow{2}{*}{Models}  & \multirow{2}{*}{FLOPs$\downarrow$} & \multicolumn{2}{c|}{I$\to$P}          & \multicolumn{2}{c|}{P$\to$I}          & \multicolumn{2}{c|}{I$\to$C}          & \multicolumn{2}{c|}{C$\to$I}          & \multicolumn{2}{c|}{C$\to$P}          & \multicolumn{2}{c|}{P$\to$C}          & \multicolumn{2}{c}{Average}           \\ \cline{3-16} 
							 &                        & Acc             & Param$\downarrow$           & Acc             & Param$\downarrow$           & Acc             & Param$\downarrow$           & Acc             & Param$\downarrow$           & Acc             & Param$\downarrow$           & Acc             & Param$\downarrow$           & Acc             & Param$\downarrow$           \\ \hline
	VGG16-base    & —                      & 71.3\%          & —               & 80.0\%          & —               & 88.5\%          & —               & 77.0\%          & —               & 61.1\%          & —               & 87.2\%          & —               & 77.5\%          & —               \\
	Two\_stage               & \multirow{3}{*}{26\%}  & 68.0\%          & 29.0\%          & 77.7\%          & 29.9\%          & 88.5\%          & 27.5\%          & 64.5\%          & 29.0\%          & 56.0\%          & 29.3\%          & 83.3\%          & 26.9\%          & 73.0\%          & 28.6\%          \\
	TCP\_w/o\_DA             &                        & 70.5\%          & 33.6\%          & 79.5\%          & \textbf{34.2\%} & 89.0\%          & 33.1\%          & 77.1\%          & 34.5\%          & 62.2\%          & 35.0\%          & 85.1\%          & 33.1\%          & 77.2\%          & 33.9\%          \\
	TCP                      &                        & \textbf{72.0\%} & \textbf{39.0\%} & \textbf{80.5\%} & 32.2\%          & \textbf{90.5\%} & \textbf{35.1\%} & \textbf{77.8\%} & \textbf{36.3\%} & \textbf{64.8\%} & \textbf{36.6\%} & \textbf{87.5\%} & \textbf{34.5\%} & \textbf{78.9\%} & \textbf{35.6\%} \\ \hline
	Two\_stage               & \multirow{3}{*}{70\%}  & 58.6\%          & 65.8\%          & 69.2\%          & 62.9\%          & 80.6\%          & 64.3\%          & 57.7\%          & 57.0\%          & 43.2\%          & 67.8\%          & 74.5\%          & 61.5\%          & 63.9\%          & 63.2\%          \\
	TCP\_w/o\_DA             &                        & 61.0\%          & 55.4\%          & 69.1\%          & 65.7\%          & 80.5\%          & \textbf{66.5\%} & 55.1\%          & 63.3\%          & 47.0\%          & 66.5\%          & 70.9\%          & 65.8\%          & 63.9\%          & 63.9\%          \\
	TCP                      &                        & \textbf{61.9\%} & \textbf{66.7\%} & \textbf{69.5\%} & \textbf{66.0\%} & \textbf{81.8\%} & 65.7\%          & \textbf{59.8\%} & \textbf{68.7\%} & \textbf{49.7\%} & \textbf{68.8\%} & \textbf{75.9\%} & \textbf{67.2\%} & \textbf{66.4\%}          & \textbf{67.1\%} \\ \hline\hline
	ResNet50-base & —                      & 74.8\%          & —               & 82.2\%          & —               & 92.3\%          & —               & 83.3\%          & —               & 70.0\%          & —               & 89.8\%          & —               & 82.1\%          & —               \\
	Two\_stage               & \multirow{3}{*}{12\%}  & 71.8\%          & 29.4\%          & 81.3\%          & 31.5\%          & 92.1\%          & 34.4\%          & 76.5\%          & 32.4\%          & 64.0\%          & 29.2\%          & 84.0\%          & 30.8\%          & 78.2\%          & 31.2\%          \\
	TCP\_w/o\_DA             &                        & 73.0\%          & 33.5\%          & 80.5\%          & 34.6\%          & 92.0\%          & 33.8\%          & 76.1\%          & 31.2\%          & 64.3\%          & 30.1\%          & 86.3\%          & 36.3\%          & 78.7\%          & 33.2\%          \\
	TCP                      &                        & \textbf{75.0\%} & \textbf{37.5\%} & \textbf{82.6\%} & \textbf{36.5\%} & \textbf{92.5\%} & \textbf{35.5\%} & \textbf{80.8\%} & \textbf{36.7\%} & \textbf{66.2\%} & \textbf{36.6\%} & \textbf{86.5\%} & \textbf{37.6\%} & \textbf{80.6\%} & \textbf{36.7\%} \\ \hline
	Two\_stage               & \multirow{3}{*}{46\%}  & 65.6\%          & 53.2\%          & 71.8\%          & 56.5\%          & 85.2\%          & 54.2\%          & 68.2\%          & 54.0\%          & 57.4\%          & 51.5\%          & 78.1\%          & 54.0\%          & 71.1\%          & 53.9\%          \\
	TCP\_w/o\_DA             &                        & 66.6\%          & 55.4\%          & 73.0\%          & 57.4\%          & 85.5\%          & 55.4\%          & 67.7\%          & 55.5\%          & 55.5\%          & 53.6\%          & 77.0\%          & 57.1\%          & 70.8\%          & 55.7\%          \\
	TCP                      &                        & \textbf{67.8\%} & \textbf{57.2\%} & \textbf{77.5\%} & \textbf{58.0\%} & \textbf{88.6\%} & \textbf{56.2\%} & \textbf{71.6\%} & \textbf{58.5\%} & \textbf{57.7\%} & \textbf{55.7\%} & \textbf{79.5\%} & \textbf{58.2\%} & \textbf{73.8\%} & \textbf{57.3\%} \\ \hline
	\end{tabular}%
	}
\end{table*}

\subsection{Implementation Details}
We mainly compare three methods: 
1) \textbf{Two\_stage}: which is the most straightforward method that applies channel pruning to the source domain task first, then fine-tune for the target domain task with the pruned model. 
2) \textbf{TCP\_w{/}o\_DA}: Our TCP method without the MMD loss, here we call it domain adaptation (DA) loss. Which also means $\beta = 0$ all the time in Eq.~(\ref{equ-loss}).
3) \textbf{TCP}: Our full TCP method with DA loss.

We evaluate all the methods on two popular backbone networks: VGG16 \cite{simonyan2014very} and ResNet50 \cite{he2016deep}. 
As baselines, VGG16-based and ResNet50-based are the original models that are not pruned. 
As for VGG16-based model, it has 13 convolutional layers and 3 fully-connected layers. We prune all the convolutional layers and the first fully-connected layer and 
we only use the activations of the second fully-connected layer as image representation and build the MMD loss which is shown in Fig.~\ref{fig-dan}. 
And as for ResNet50-based model, we use similar settings as VGG16-based model with a few differences. 
Because of the shortcut and residual branch structure, we only prune the inside convolutional layers of each bottleneck block. The MMD loss is built with the only fully-connected layer. 
Moreover, we also take the Batch Normalization (BN)~\cite{ioffe2015batch} layers into consideration and reconstruct the whole model during pruning.

In practice, all the input images are cropped to a fixed size $224\times224$ and randomly sampled from the resized image with horizontal flip and mean-std normalization. 
At first, we fine-tune all the UDA models on each unsupervised domain adaptation tasks for 200 epochs with learning rate 
from 0.01 to 0.0001 and the batch size $=$ 32. During pruning, we set $\mathcal{K}=64$ which means 64 channels will be removed after each pruning iteration. 
After that, extra 5 epochs are adopted to help the pruned model to converge. And we follow \cite{hu2018novel} to prune the baseline with different compression rate and make the compression rate as different as possible. 
The VGG16-based baseline is pruned with 26\% and 70\% FLOPs reduced while the ResNet50-based baseline is pruned with 12\% and 46\% FLOPs reduced. 
ResNet50 has lower compression rate since the bottleneck structure stops some layers from being pruned.

We follow standard evaluation protocol for UDA and use all source examples with labels and all target examples without labels \cite{gong2013connecting}. 
The labels for the target domain are only used for evaluation. 
We adopt \textit{classification accuracy} on the target domain and \textit{parameter reduction} as the evaluation metrics:~higher accuracy and fewer parameters indicate better performance.

\subsection{Results and Analysis}
Firstly, we evaluate all the tasks on Office-31 dataset. The results are shown in TABLE~\ref{tb-office}. As can be seen, our TCP method outperforms other methods under the same compression rate (FLOPs reduction) and can reduce more parameters. 
And the FLOPs in convolutional layers is calculated by:
\begin{equation}
	\label{equ-flops}
	FLOPs~=~HWC_{in}K^{2}C_{out},
\end{equation} 
where $H, W, C_{out}$ is the height, width and channel number of output feature map, $K$ is the kernel size, $C_{in}$ refers to the number of input channels, and the bias item is ignored due to its small contribution. Moreover, 
It is important and interesting that TCP achieves even better performance than the baseline model (which is not pruned).
This is probably because some redundant channels in the base model are removed thus negative transfer is reduced. Especially for the results of ResNet50-based models, our baseline is almost the same as the result of DAN in~\cite{long2015learning}.
However, we can get better performance on half of the tasks and we even get 100\% on task W$\to$D after 46\% FLOPs have been reduced.

Secondly, we evaluate our methods on ImageCLEF-DA dataset and the results are shown in TABLE~\ref{tb-clef}. We can draw the same conclusion that TCP performs better 
on all tasks on ImageCLEF-DA dataset. We get higher accuracy than the baseline on all the VGG16-based experiments after 26\% FLOPs have been reduced, 
and we also get higher accuracy on the target dataset on half of the tasks on ResNet50-based experiments after 12\% FLOPs have been reduced, compared with the baseline which is almost the same as DAN~\cite{long2015learning}.

\begin{figure}[t]
	\centering
	\subfigure[VGG16-based]{
		\centering
		\includegraphics[scale=0.16]{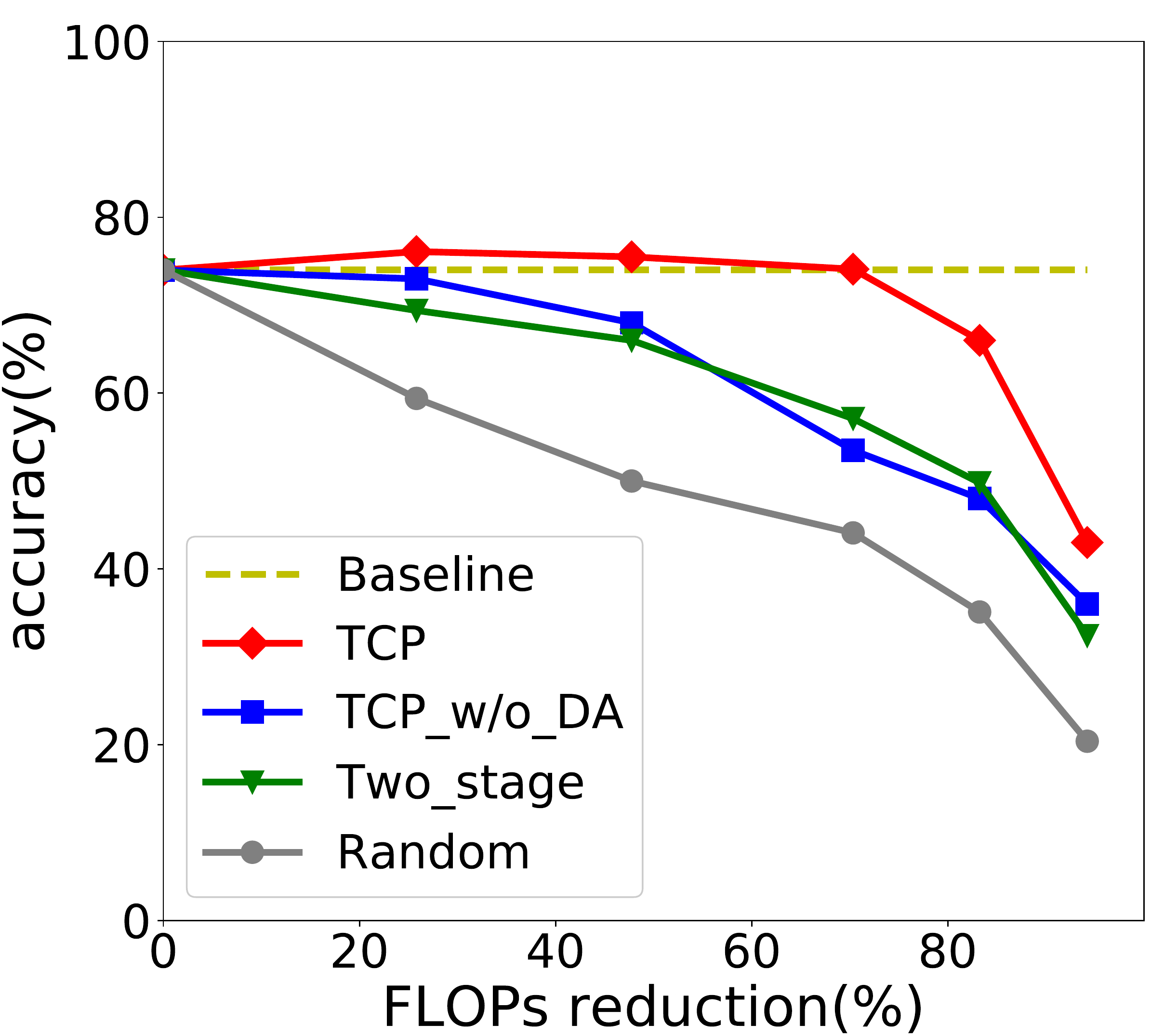}
		\label{fig-sub-vgg16}}
	\subfigure[ResNet50-based]{
		\centering
		\includegraphics[scale=0.16]{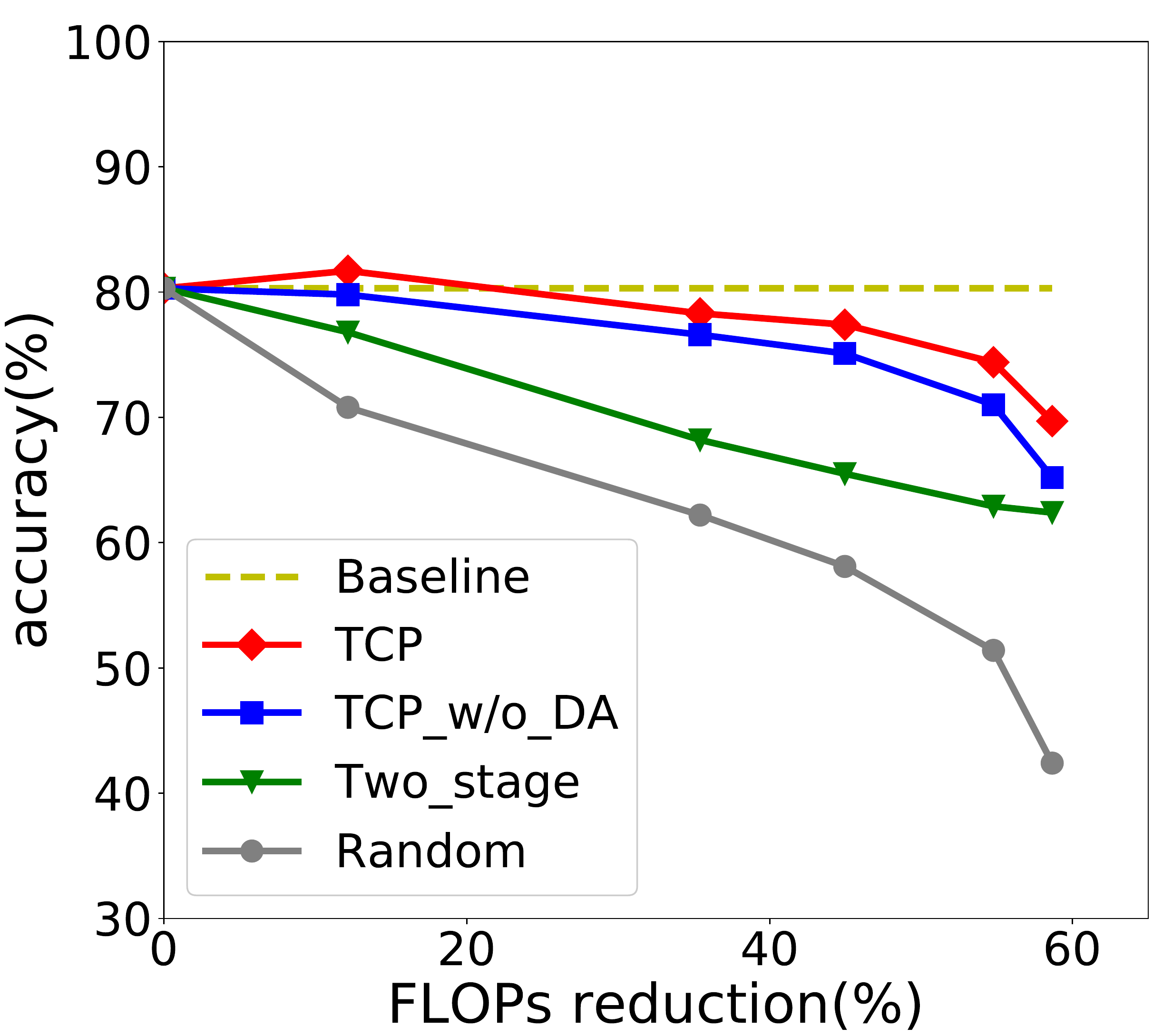}
		\label{fig-sub-res}}
	\vspace{-.1in}
	\caption{The pruning result on task A$\to$W with more compression rate.}
	\label{fig-office_vggandres}
	\vspace{-.1in}
\end{figure}

Apart from TABLE~\ref{tb-office} and TABLE~\ref{tb-clef}, Fig.~\ref{fig-office_vggandres} shows the comparison for all methods. And we also add a 
Random method which randomly removes a certain number of channels to achieve the same reduction of FLOPs. Combining these results, more conclusions can be made. 1) Compared with Two\_stage, 
TCP is more efficient because it is a unified framework and treat the pruning as a single optimization problem, while Two\_stage is a split method and it 
does not take the target domain into consideration while pruning. 2) Compared with TCP\_w{/}o\_DA, the full TCP uses the transfer channel evaluation to represent the discrepancy between the source and target domains. 
We try to remove those less important channels for both source and target domains and reduce negative transfer by reducing domain discrepancy. 3) As can be seen from Fig.~\ref{fig-office_vggandres}, our TCP outperforms other methods on unlabeled target 
dataset under different compression rate. 4) This indicates that TCP is generic, accurate, and efficient, which can dramatically reduce the computational cost of a deep UDA model without sacrificing the performance.

\begin{figure}[t]
	\centering
	\subfigure[Baseline:~$Source$ = A]{
		\centering
		\includegraphics[scale=0.17]{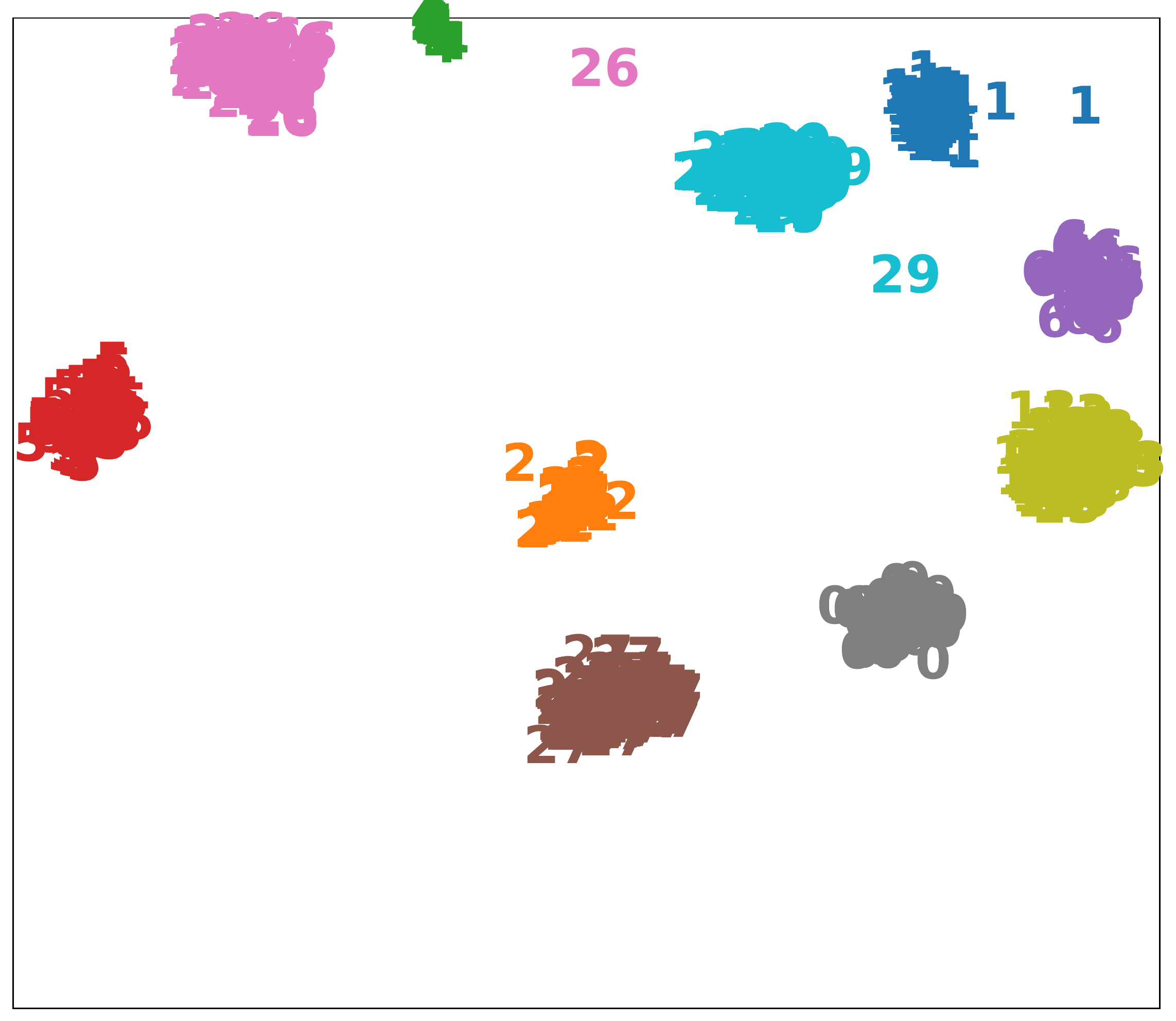}
		\label{fig-sub-awressource}}
	\vspace{-.1in}
	\subfigure[Two\_stage:~$Target$ = W]{
		\centering
		\includegraphics[scale=0.17]{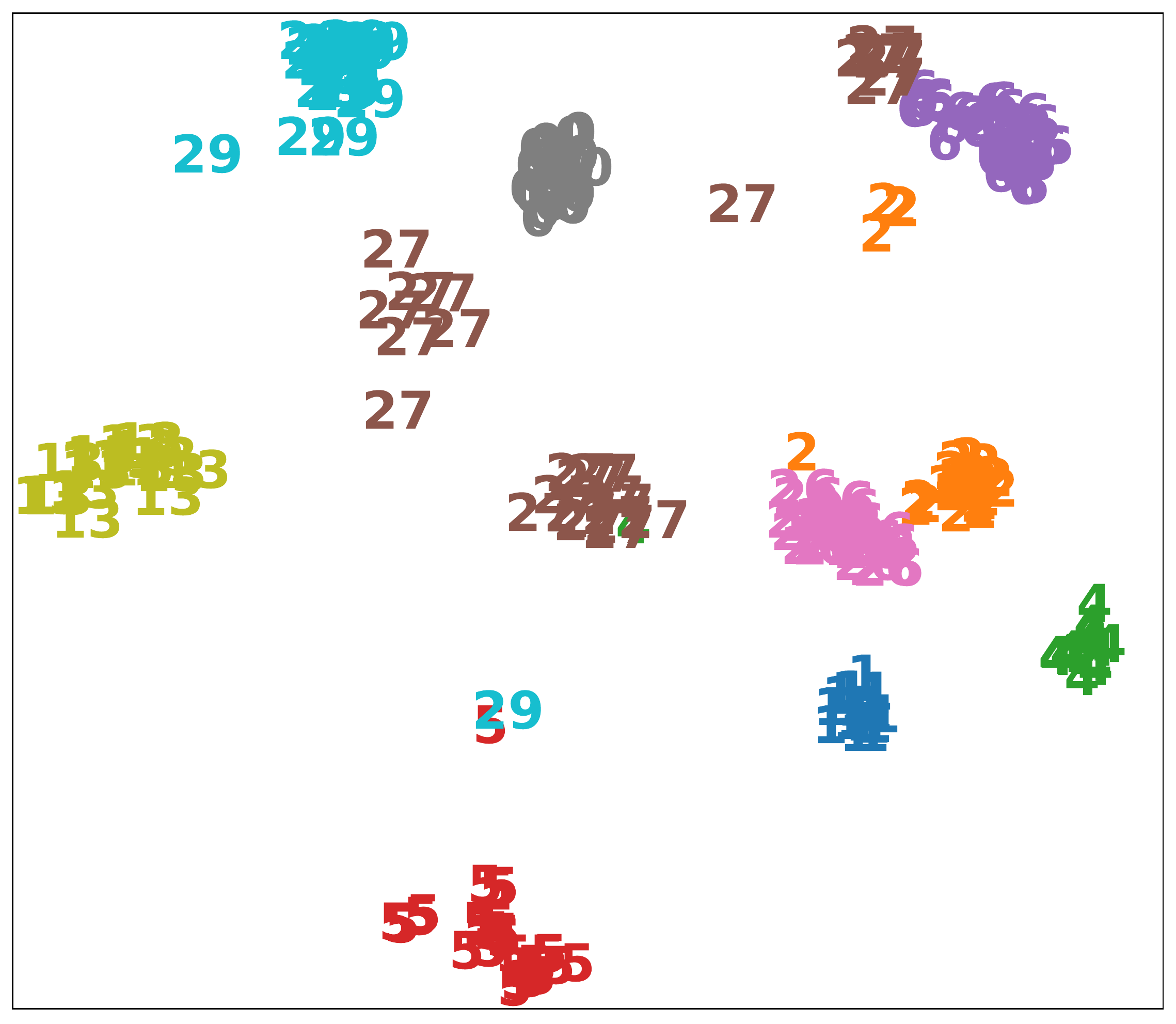}
		\label{fig-sub-awresprandft}}
	\vspace{-.1in}
	\subfigure[TCP\_w{/}o\_DA:~$Target$ = W]{
		\centering
		\includegraphics[scale=0.17]{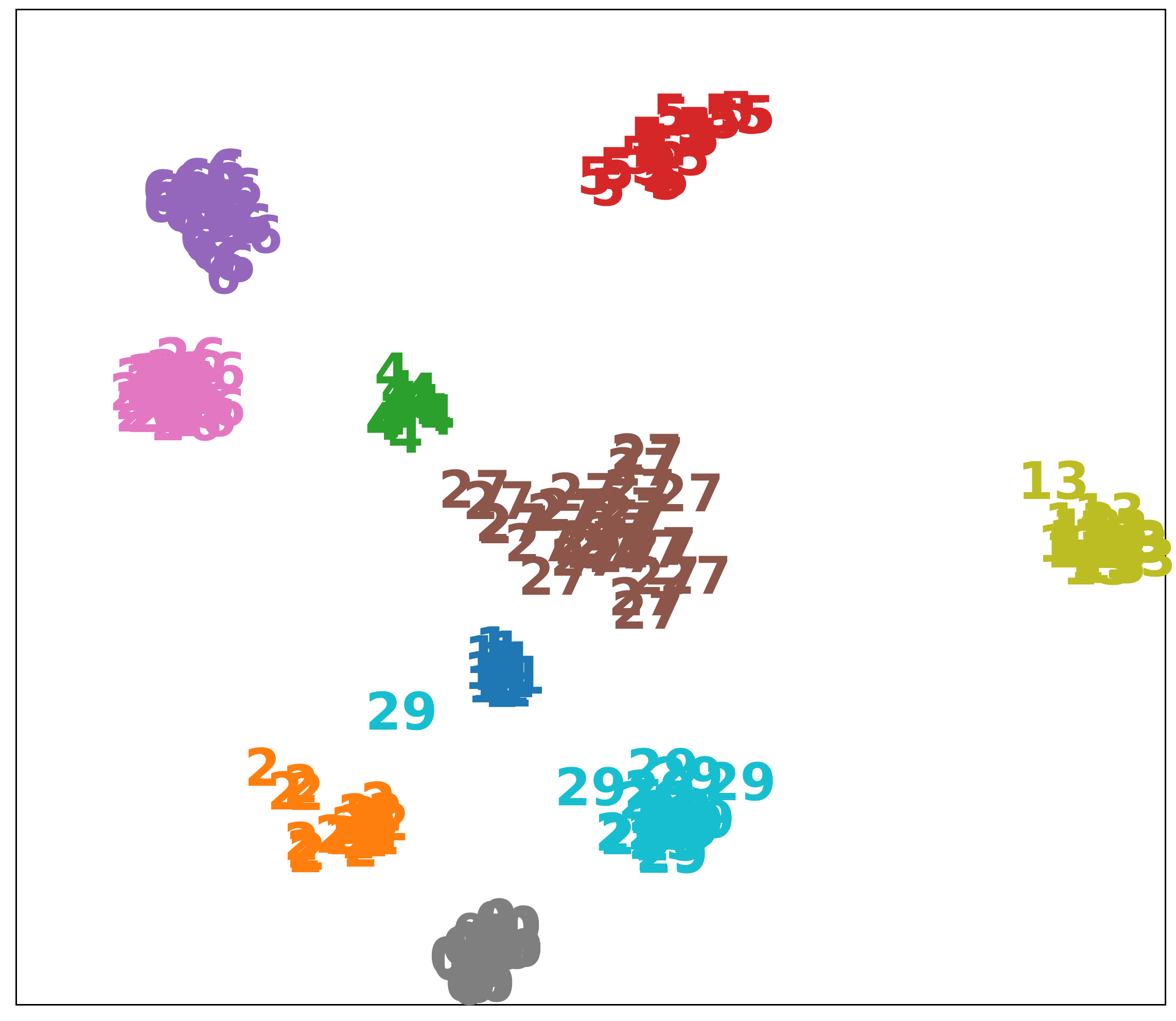}
		\label{fig-sub-awrescls}}
	\subfigure[TCP:~$Target$ = W]{
		\centering
		\includegraphics[scale=0.17]{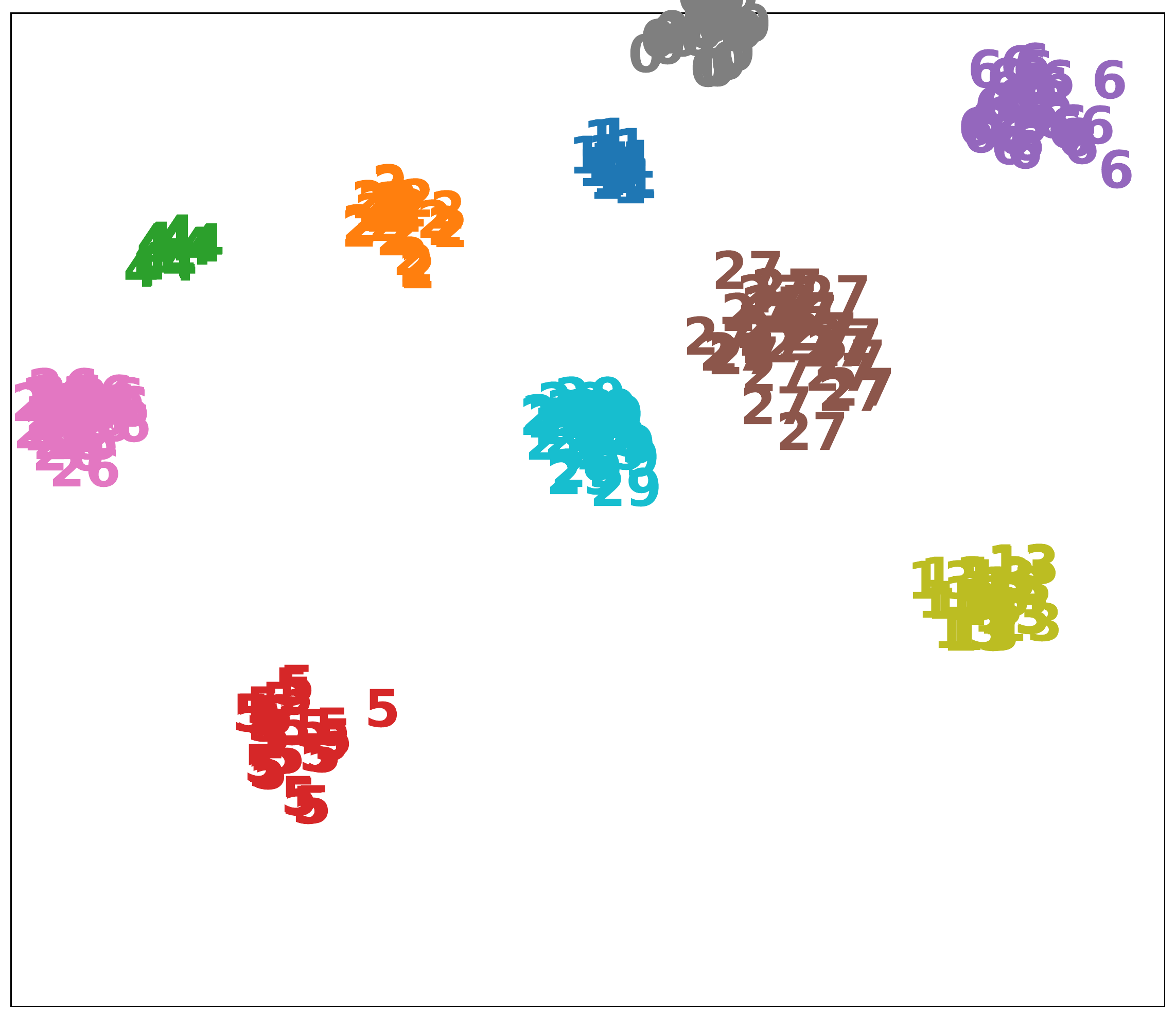}
		\label{fig-sub-awresclsmmd}}
	\caption{The t-SNE visualization of network activations. (a) is generated by ResNet50-based baseline without pruning on source domain. (b)(c)(d) are generated by 
	ResNet50-based (with 12\% FLOPs pruned) with our three methods on target domain respectively. Best view in color.}
	\label{fig-awresinfo}
	\vspace{-.15in}
\end{figure}

\subsection{Effectiveness Analysis}
\textit{Visualization analysis}. 
To evaluate the effectiveness of TCP in reducing negative transfer, in Fig.~\ref{fig-awresinfo}, we follow \cite{donahue2014decaf} to visualize the model activations of task A$\to$W pruned by different methods using t-SNE~\cite{donahue2014decaf}.
Fig.~\ref{fig-sub-awressource} shows the results of ResNet50-based baseline without pruning on the source domain. And Fig.~\ref{fig-sub-awresprandft}, Fig.~\ref{fig-sub-awrescls} and Fig.~\ref{fig-sub-awresclsmmd} 
denote the result of ResNet50-based models on the target domain, which have been pruned by 12\% FLOPs with our three methods Two\_stage, TCP\_w{/}o\_DA and TCP respectively. 
The colored digits represent the ground truth of the examples, so the number is from 0 to 30, which denotes target dataset has 31 categories. Here we randomly pick 10 categories to visualize.
As can be seen, the target categories are discriminated more clearly with the model pruned by our TCP method. This suggests that our TCP method is effective in learning more transferable features by reducing the cross-domain divergence.

\begin{figure}[t!]
	\centering
	\subfigure[Office31:~A$\to$W]{
		\centering
		\includegraphics[scale=0.41]{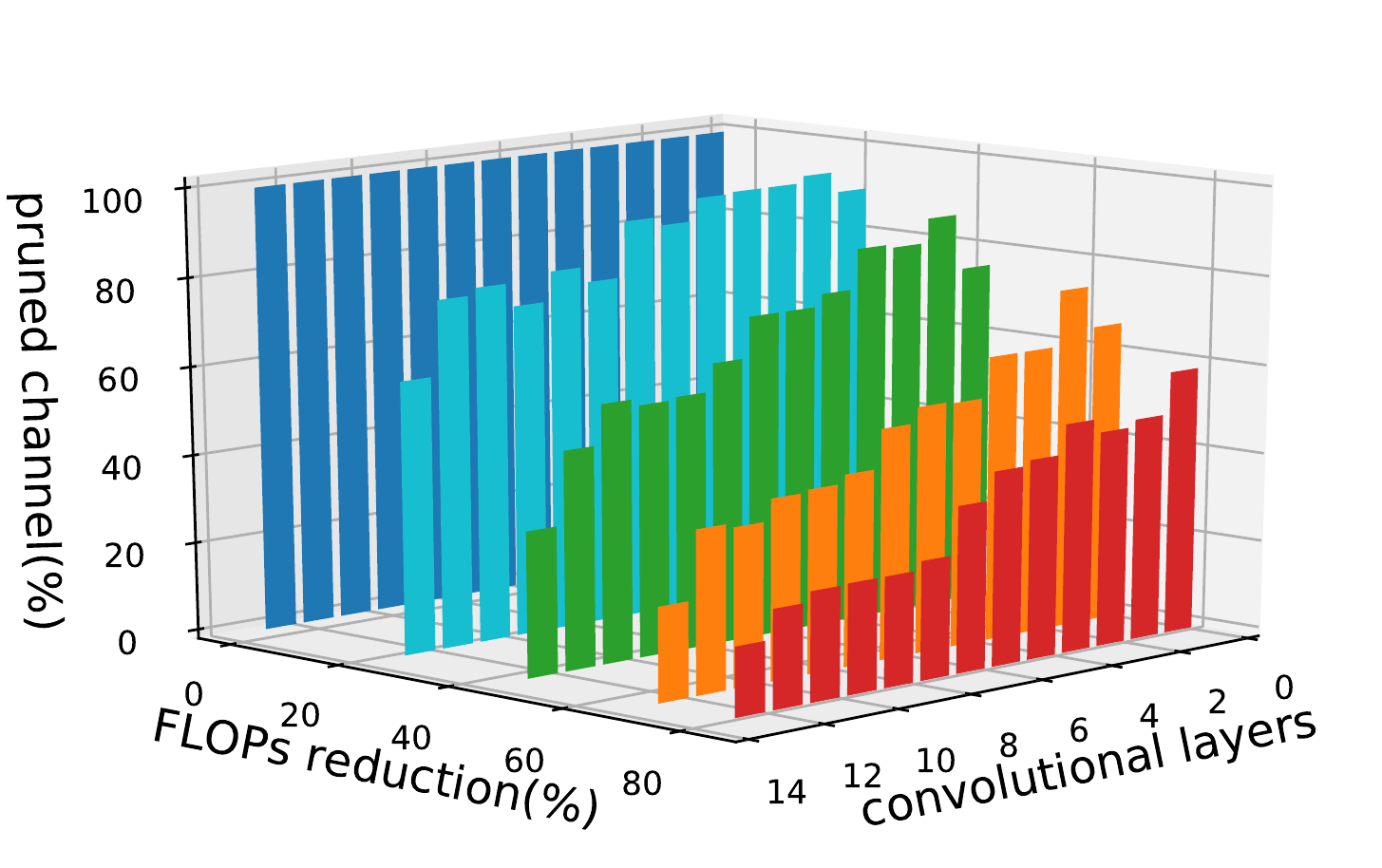}
		\label{fig-sub-awstruc}}
	\subfigure[ImageCLEF-DA:~I$\to$P]{
		\centering
		\includegraphics[scale=0.41]{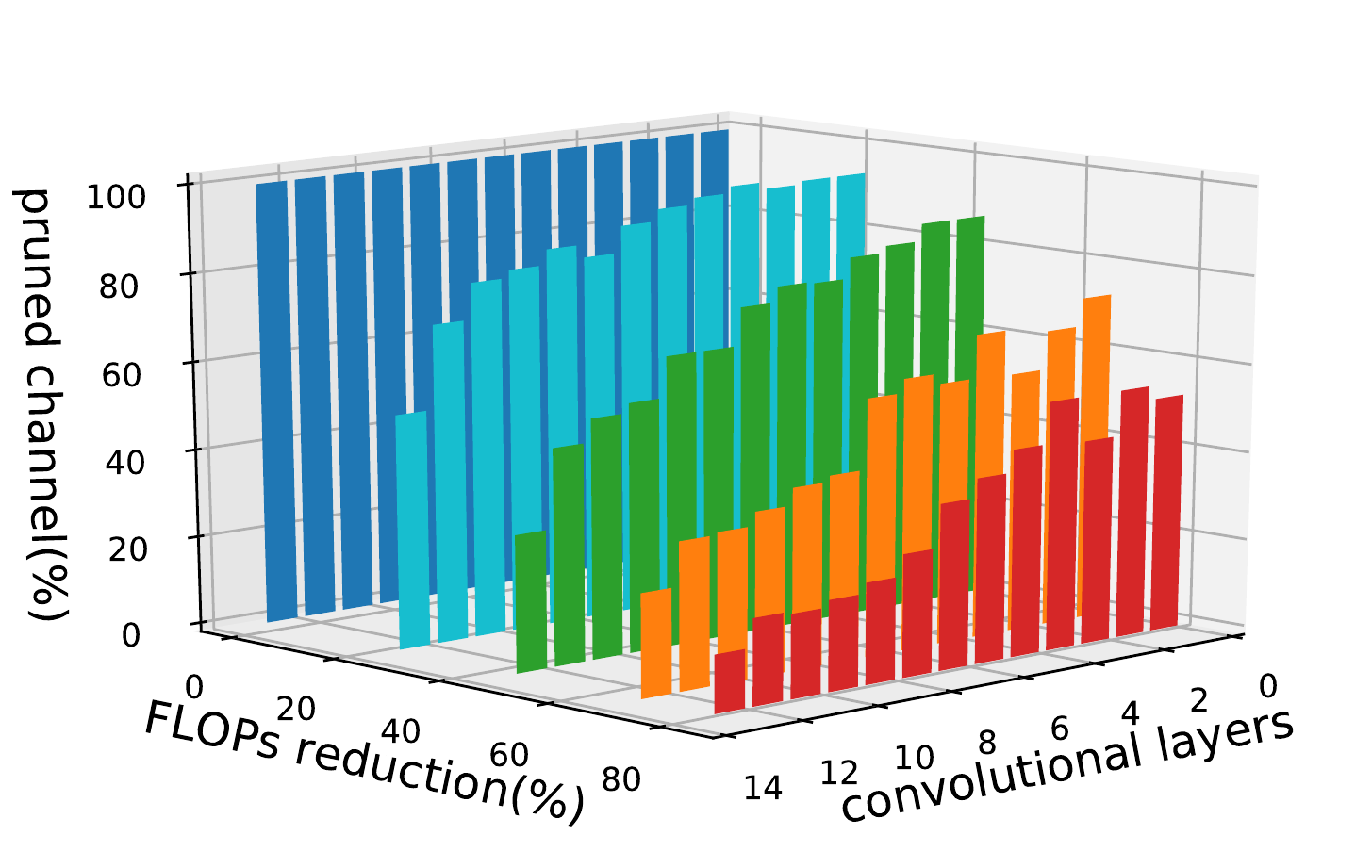}
		\label{fig-sub-ipstruc}}
	\vspace{-.1in}
	\caption{The pruned structure of all the 13 convolutional layers of VGG16-based network on different dataset for deep unsupervised domain adaptation. Best view in color.}
	\label{fig-awandipstruc}
	\vspace{-.2in}
\end{figure} 

\textit{Pruned structure analysis}. 
To explore if there is any pattern in the structure of the pruned models, we show the structure of pruned models on task A$\to$W and I$\to$P in Fig.~\ref{fig-awandipstruc} with TCP. As we can see, higher layers have more redundancy than lower layers in VGG16-based models, and our TCP prefer pruning the higher layers. This is reasonable for UDA because the lower layers usually 
encode common and important features for both source and target domains. 
Moreover, because there are more parameters in higher layers of CNN, especially the first fully-connected layer in VGG16, our TCP thus can prune more parameters under almost the same compression rate. 
The same result can be observed on ResNet50-based models.

\section{Conclusion and Future Work}
\label{sec-conclu}
In this paper, we propose a unified Transfer Channel Pruning~(TCP) approach for accelerating deep unsupervised domain adaptation models. TCP is capable of compressing the deep UDA model by pruning less important channels while simultaneously learning transferable features by reducing the cross-domain distribution divergence. Therefore, it reduces the impact of negative transfer and maintains competitive performance on the target task. TCP is a generic, accurate, and efficient compression method that can be easily implemented by most deep learning libraries. Experiments on two public benchmark datasets demonstrate the significant superiority of our TCP method over other methods.

In the future, we plan to extend TCP in pruning adversarial networks and apply it to heterogeneous UDA problems.

\section{Acknowledgment}
This work is supported in part by National Key R \& D Plan of China~(No.2017YFB1002800), NSFC~(No.61572471), and Beijing Municipal Science \& Technology Commission~(No.Z171100000117017).

\bibliographystyle{IEEEtran}
\bibliography{ijcnn19}

\end{document}